\documentclass[10pt,twocolumn,letterpaper]{article}

\usepackage[pagenumbers]{cvpr} % To force page numbers, e.g. for an arXiv version

\usepackage{graphicx}
\usepackage{amsmath}
\usepackage{amssymb}
\usepackage{booktabs}
\usepackage{comment}
\usepackage{times}
\usepackage{epsfig}
\usepackage{graphicx}
\usepackage{amsmath}
\usepackage{amssymb}
\usepackage{algorithm}
\usepackage{algorithmicx}
\usepackage{algpseudocode}
\usepackage[usestackEOL]{stackengine}
\usepackage{adjustbox}
\usepackage{multirow}
\usepackage{mathtools}
\usepackage{makecell}
\usepackage{tabularx}
\usepackage{xcolor}

\DeclarePairedDelimiter\floor{\lfloor}{\rfloor}

\newcommand\eatpunct[1]{}

\newcolumntype{Y}{>{\centering\arraybackslash}X}
\usepackage{flushend}

\usepackage[pagebackref,breaklinks,colorlinks]{hyperref}

\usepackage[capitalize]{cleveref}
\crefname{section}{Sec.}{Secs.}
\Crefname{section}{Section}{Sections}
\Crefname{table}{Table}{Tables}
\crefname{table}{Tab.}{Tabs.}

\def\cvprPaperID{3337} % *** Enter the CVPR Paper ID here
\def\confName{CVPR}
\def\confYear{2023}

\begin{document}

\title{
VideoCoCa: Video-Text Modeling with Zero-Shot Transfer \\ from Contrastive Captioners
}

\author{Shen Yan$^\dag$\thanks{
Equal contribution.} \;\; Tao Zhu$^\dag$$^*$\;\; Zirui Wang$^\dag$ \;\;  Yuan Cao$^\dag$ \;\; Mi Zhang$^\ddag$\;\; \\ 
Soham Ghosh$^\dag$ \;\; Yonghui Wu$^\dag$\;\; Jiahui Yu$^\dag$ \\
$^\dag$Google Research \quad
$^\ddag$The Ohio State University \\
{\tt \small \{shenyan, tzhu, ziruiw, yuancao, jiahuiyu\}@google.com \;\; \;\;}\\
}

\maketitle

%%%%%%%%% ABSTRACT
\begin{abstract}
We explore an efficient approach to establish a foundational video-text model. We present VideoCoCa that maximally reuses a pretrained image-text contrastive captioner (CoCa) model and adapt it to video-text tasks with minimal extra training.
While previous works adapt image-text models with various cross-frame fusion modules, we find that the generative attentional pooling and contrastive attentional pooling layers in CoCa are instantly adaptable to flattened frame embeddings, yielding state-of-the-art results on zero-shot video classification and zero-shot text-to-video retrieval. Furthermore, we explore lightweight finetuning on top of VideoCoCa, and achieve strong results on video question-answering and video captioning.

\end{abstract}

\section{Introduction}

%Foundation models \cite{bommasani2021opportunities} for vision-language pretraining (VLP) have been shown as a promising direction for multimodal learning.
%The goal is to pretrain models on large-scale data sources and then transfer efficiently to diverse downstream tasks.
Recently large-scale pretrained foundation models~\cite{bommasani2021opportunities} have achieved successes across a wide range of tasks on natural language processing and multimodal problems, with versatility to transfer on a diverse set of downstream tasks.
For image-text models, pioneering works have achieved strong performance by utilizing noisy web-crawled image-text pairs,
such as CLIP\cite{radford21clip} and ALIGN\cite{jia2021scaling} for crossmodal retrieval and SimVLM\cite{wang2021simvlm} for multimodal understanding.
More recent works \cite{yuan2021florence, yu2022coca, wang2022git, wang2022image, wang2022omnivl} further improved the pretraining approaches to enable better and more downstream vision-language tasks.
In particular, Contrastive Captioners (CoCa)\cite{yu2022coca} is directly applicable to classification, crossmodal retrieval and multimodal understanding through a unified contrastive and captioning learning schemes,
achieving superior performance on vision and vision-language tasks including image classification, image-text retrieval, and visual question answering.

\begin{figure*}[t]
    %\vspace{-\baselineskip}
	\centering
	\hspace*{0.8cm}   
    \includegraphics[width=0.89\textwidth]{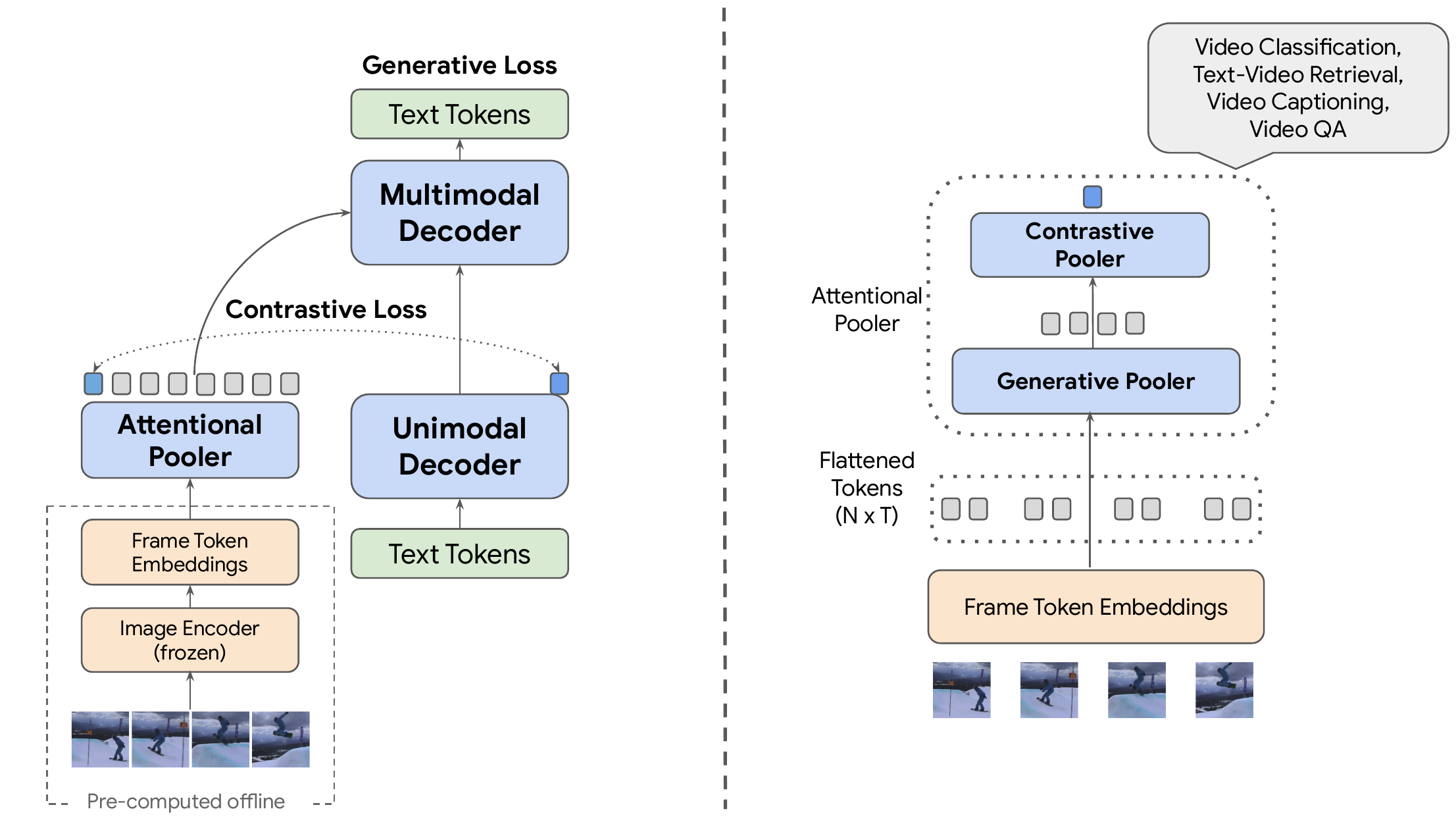}
    \caption{\textbf{Left}: Overview of VideoCoCa. All weights of the pretrained CoCa model are reused, without the need of learning new modules. We compute frame token embeddings offline from the frozen CoCa image encoder. These tokens are then processed by a generative pooler and a contrastive pooler on all flattened frame tokens, yielding a strong zero-shot transfer video-text baseline. When continued pretraining on video-text data, the image encoder is frozen, while the attentional poolers and text decoders are jointly optimized with the contrastive loss and captioning loss, thereby saving heavy computation on frame embedding. \textbf{Right}: An illustration of the attentional poolers and flattened frame token embeddings. We flatten $N \times T$ token embeddings as a long sequence of frozen video representations.
    }
    \vspace{-\baselineskip}
	\label{fig:model-overview}
\end{figure*}
%The weights of all components are directly loaded from the pretrained image-text CoCa model

%In this paper, we question whether any heavy video adaptor is needed and explore the \textit{limit} of transferring image-text foundation models for video-text modelling.  
To extend image-text models to video, recent work has focused on either the visual domain by aggregating temporal information with new modules~\cite{wang2021actionclip,alayrac2022flamingo,ni2022XCLIP}, or on the textual domain by proposing different sampling methods to curate higher-relevance text-video pairs~\cite{yang2021taco,xu2021videoclip} for efficient contrastive training. Given a well-pretrained image-text foundation model, it is natural to question whether any heavy video-specific adaptor or many video-specific data is needed when transferring to video-text modelling

%it can be efficiently adapted to model video-text without the need of learning many video-specific parameters or accessing many video-specific data.

%Given a well-pretrained image-text foundation model, it is natural to question whether it can be efficiently adapted to modeling video-text without much extra training. 

In this paper, we explore an efficient approach to establish a foundational video-text model for tasks including open-vocabulary video classification, text-to-video retrieval, video captioning and video question-answering. We present VideoCoCa, a minimalist approach that extends the image-text contrastive captioners (CoCa) \cite{yu2022coca} to video-text tasks. The design principle of VideoCoCa is to maximally reuse a pretrained image-text CoCa model and minimize additional training overhead. 
%We therefore naïvely feed video inputs to the pretrained CoCa image model as a sequence of independent image frames, flatten token embeddings as a long sequence of frozen video representation and apply the attentional poolers on top. 
Figure~\ref{fig:model-overview} (left) shows an overview of our proposed model. As illustrated, VideoCoCa reuses the architecture and pretrained model weights of CoCa and applies the image encoder to a sequence of sampled frames of a video independently. The outputs are flattened and concatenated to form a long sequence of frozen token embeddings, with the same attentional poolers applied on top of them.

%, and simply treats each video example as a sequence of independent image frames to obtain the flattened token embeddings as a long sequence of frozen video representation, while the same attentional poolers are applied on top.

Without any additional video-specific parameter or early temporal fusion, and without accessing any video or video-text data, VideoCoCa consistently outperforms existing video-text models on video recognition tasks. It achieves new state-of-the-art results on zero-shot video classification on Kinetics 400/600/700~\cite{kay_arxiv_2017}, UCF101~\cite{soomro2012ucf101}, HMDB51~\cite{kuehne2011hmdb}, and Charades~\cite{sigurdsson2016hollywood}, as well as zero-shot text-to-video retrieval on MSR-VTT~\cite{xu2016msrvtt}, ActivityNet Captions~\cite{krishna2017densecaptioning} and VATEX~\cite{wang2019vatex}.
We also explore different lightweight finetuning approaches on video-text data to further boost results and achieve the best quality-efficiency trade-offs. VideoCoCa is able to obtain further improvement on top of its zero-shot transferred model architecture and achieves strong results on the tasks of video question answering (iVQA~\cite{yang2020just}, MSRVTT-QA~\cite{xu2017video}, MSVD-QA~\cite{xu2017video}, ActivityNet-QA~\cite{yu2019activityqa}) and video captioning (MSR-VTT~\cite{xu2016msrvtt}, ActivityNet~\cite{krishna2017densecaptioning}, VATEX~\cite{wang2019vatex}, Youcook2~\cite{zhou2017automatic}). %Our approach establishes a simple and strong video-text baseline for future research on leveraging pretrained image-text foundation models for video-text tasks.

\section{Related Work}
\paragraph{Efficient video learners from image-text models.}
CLIP~\cite{radford21clip} and ALIGN~\cite{jia2021scaling} were the first two works to gain popularity on zero-shot image classification and crossmodal alignment tasks, by training dual-encoder models with contrastive loss on noisy image-text pairs. Since then, many works~\cite{Luo2021CLIP4Clip, wang2021actionclip, ju2021prompting, lin2022frozen, xue2022clipvip} have reported that retraining a new model for video-text models is extremely challenging due to efficiency considerations, and they have investigated ways to transfer the knowledge from the image-text CLIP model to end-to-end action recognition and video-text retrieval. More recently, there have been works that show efficient ways of adapting strong image-text models to video foundation models with minimal extra computation. For example, X-CLIP~\cite{ni2022XCLIP} uses a cross-frame attention and video-specific prompting for video classification. CLIP-Hitchhiker~\cite{bain2022cliphitchhikers} aggregates video features by using the weighted mean of image embeddings with text as similarity queries, achieving strong results on text-to-video retrieval. In line with these research, we investigate efficient adaptation methods for zero-shot transfer in the video-text domain by adapting attentional poolers.

\vspace{-1\baselineskip}
\paragraph{Adapting image transformers to videos.}
The success of these efficient video learners is largely due to the fact that they initialize their video models using pretrained image models. Early works~\cite{carreira2017quo,NonLocal2018,xie2017rethinking} on 3D CNNs have demonstrated that image classification models can be inflated into spatio-temporal feature extractors and their pretrained weights provide a valuable initialization. For Transformers, several prior works~\cite{bertasius2021spacetime,arnab2021vivit,yan2022multiview,liu2021video,li2022uniformer} illustrated the effectiveness of this approach for video recognition. A common idea among these works is the adaptation of the ViT~\cite{dosovitskiy2020image} architecture for videos by extending self-attention from 2D image space to 3D spatio-temporal volume. For example, TimeSformer~\cite{bertasius2021spacetime} uses divided space-time attention, where temporal and spatial attention layers in each block are initialized with the same weights from the corresponding attention layer in ViT. ViViT~\cite{arnab2021vivit} initializes the weights of the tubelet embedding using the central frame initialization from the image model, where all temporal positions are initialized with zeros except at the center position. MTV~\cite{yan2022multiview} further extends this idea by initializing each view of the multiview encoder. Building on these research, we explore the use of all component weights of CoCa for the model initializations. 

\vspace{-1\baselineskip}
\paragraph{Zero-shot transfer for video understanding.}
LiT~\cite{zhai2021lit} shows that a locked pre-trained image model with an unlocked text model is the most efficient to turn existing vision backbones into zero-shot learners. Flamingo~\cite{alayrac2022flamingo} freezes the weights of the pretrained vision and language models, and only trains the cross-attention layers and perceiver resampler~\cite{jaegle2021perceiver} for efficient few-shot learning on video understanding. FrozenBiLM~\cite{yang2022zeroshot} improves zero-shot videoQA by learning additional lightweight adapter modules of the frozen bidirectional language models. Socratic models~\cite{zeng2022socratic} enable zero-shot multimodal-informed prompting using multiple visual-audio-language pretrained models even without requiring finetuning. Similar to the perceiver resampler~\cite{jaegle2021perceiver} used in Flamingo~\cite{alayrac2022flamingo} and the query learner used in PaLI\cite{chen2022pali}, we adopt the generative pooler in CoCa as the downsampler to learn a fixed set of spacetime tokens. It does not introduce new parameters but still allows for lightweight adaptation.

\section{VideoCoCa} \label{sec:method}
%This section begins with an overview of image-text contrastive captioners, CoCa~\cite{yu2022coca} in Section~\ref{sec:method_prelim}. We then discuss its lightweight adaptation to video-text models in Section~\ref{sec:method_videococa}. Finally, we explore different model finetuning choices in Section~\ref{sec:method_tuning}.  

%We now describe the VideoCoCa approach as a derivative of CoCa.

\subsection{Preliminaries: CoCa} \label{sec:method_prelim}
Contrastive Captioners (CoCa)~\cite{yu2022coca} is an encoder-decoder architecture that combines contrastive pretraining approaches like CLIP~\cite{radford21clip} with generative pretraining apporaches like SimVLM~\cite{wang2021simvlm}. %It is mainly used to learn image-level, text-level, and joint image-text multimodal representations. 
It is designed to facilitate image, text and image-text representation learning. CoCa adopts a cascaded decoder design, where the bottom half unimodal decoder encodes the text context with causally masked self-attention, and the top half multimodal decoder uses cross-attention to align image and text. The model is trained with joint contrastive loss and captioning loss. For text representations, the [CLS] token from the unimodal decoder is used as a global text representation for the contrastive loss, and the captioning loss is applied per text token to learn fine-grained visual-textual information.

% For image representations, CoCa adopts task-specific attentional poolers~\cite{lee2018set} to customize image representations to be used for different training objectives and downstream tasks. The attentional pooler is referred as a single multi-head attention layer with $n_{query}$ learnable queries, with the vision transformer encoder output as both keys and values. With the attentional poolers, the model learns to pool image representations with different lengths for both training objectives, allowing different needs for different tasks. In practice, CoCa uses $n_{query} = 256$ in pretraining for the generative loss, and $n_{query} = 1$ for the contrastive loss. In our work, we adopt the cascade pooler design, where the contrastive pooler is applied on top of the outputs of the generative pooler.

CoCa uses two attentional pooling layers ~\cite{lee2018set} (poolers in short) to extract image representations, in which a $n_{query} = 256$ generative attentional pooling layer is used to generate embeddings for captioning loss, and a $n_{query} = 1$ contrastive attentional pooling layer together with its outputs as contrastive image embedding. The benefit of such an architecture design is two-fold. First, the pooler yields a fixed number of tokens (for example 256 tokens as generative embedding and 1 token as contrastive embedding) regardless of input image resolutions, making the image encoder and the text decoder more modular and adaptable to other modalities. Second, the pooler serves as a lightweight adaptor so that the pretrained ViT as backbone remains frozen for many downstream tasks. For example, it is shown that by only finetuning the poolers, a pretrained CoCa can already achieve 90.6\% top-1 ImageNet accuracy, in which case the frozen ViT did not see any ImageNet data. In this work, we adopt this design and explore to adapt to video-text domain.

%In this work, we further adopts this design principle to adapt image-text CoCa to video-text modeling. We use the default cascaded pooler architecture~\cite{yu2022coca}, where the contrastive pooler is applied on top of the outputs of the generative pooler.

\subsection{Transferring CoCa to Video-Text Tasks}
\label{sec:method_videococa}

We illustrate how image CoCa can be quickly turned into VideoCoCa by tuning a small portion of parameters. We denote our input minibatch of videos as $\mathbf{V} \in \mathbb{R}^{B \times T \times H \times W \times C}$, where \(T\) is the number of frames uniformly sampled from a video. We follow ViT~\cite{dosovitskiy2020image} and extract tokens from images by partitioning an image into non-overlapping patches and linearly projecting them. All tokens are then concatenated together to form a sequence, resulting in a minibatch of the sequence tokens $\mathbf{\Tilde{z}}$ of shape $\mathbb{R}^{(B \times T) \times N \times d}$, where $N = \floor{\frac{H}{h}} \times \floor{\frac{W}{w}}$. A positional embedding $\mathbf{p} \in \mathbb{R}^{N \times d}$ is next added to this sequence to obtain $\mathbf{z}$. We follow~\cite{yu2022coca} to extract the text representations (see Section~\ref{sec:method_prelim}). The frame-level representations $\mathbf{z}^{L} \in \mathbb{R}^{(B \times T) \times N \times d}$ are obtained by forwarding $\mathbf{z}$ into the CoCa image encoder, where $L$ is the number of encoder layers. We consider the following alternatives to adapt CoCa to videos.

\vspace{-2mm}
\paragraph{Attentional poolers.}
\label{sec:method_videococa_pooler}
To obtain video representations, we concatenate all spatial tokens together along the temporal dimension into $\mathbf{z}^{L} \in \mathbb{R}^{B \times (T \times N) \times d}$, which is then fed into the generative and the contrastive poolers. See Figure \ref{fig:model-overview} (right) for a detailed illustration. This model corresponds to a late fusion of temporal information, similar to the factorized encoder. Compared to ~\cite{yu2022coca}, where additional new poolers are added on top of the frame-level representations to learn video representations, this model does not add any novel learnable layers, allowing reuse of all parameters from the pretrained CoCa model with minimal extra computation, thereby enabling zero-shot transfer to video-text tasks from image-text models. 

\vspace{-2mm}
\paragraph{Factorized encoder.}
\label{sec:method_videococa_factorized_encoder}
This model additionally adds a Transformer encoder on top of the contrastive pooler. The frame-level representations $\mathbf{z}^{L}$, is first fed into the generative pooler and the contrastive pooler to get the spatial embeddings $\mathbf{z}^{L_s} \in \mathbb{R}^{(B \times T) \times d}$. The spatial embeddings $\mathbf{z}^{L_s}$ are then reshaped to into $\mathbf{z}^{L_s} \in \mathbb{R}^{B \times T \times d}$ and fed into a transformer encoder consisting of $L_t$ layers to model interactions between tokens from different frames. The output tokens are used for the captioning loss and their global average pooled embeddings over the temporal dimension are used for the contrastive loss. Following~\cite{arnab2021vivit, yan2022multiview}, we choose $L_t = 4$. Note that different from~\cite{arnab2021vivit,yan2022multiview}, where the spatial embeddings are summarized by the prepended learnable class token $\mathbf{z}_{cls} \in \mathbb{R}^{d}$~\cite{devlin_naacl_2019}, we use the output of the contrastive pooler as representation.

\vspace{-2mm}
\paragraph{Joint space-time encoder.}
\label{sec:method_videococa_Unfactorized_encoder}
This model is commonly known as spatio-temporal attention model~\cite{arnab2021vivit} or joint space-time model~\cite{bertasius2021spacetime}. Specifically, we reshape the sequence tokens into $ \mathbf{\Tilde{z}} \in \mathbb{R}^{B \times (T \times N) \times d}$ and then add the positional embedding $\mathbf{p} \in \mathbb{R}^{(T \times N) \times d}$ to obtain $\mathbf{z}$. $\mathbf{p}$ is initialized by temporally repeating the positional embedding from the pretrained image model. This allows CoCa image encoder to encode pairwise interactions between all spatial-temporal tokens from the first layer. The spatial-temporal representations $\mathbf{z}^{L}$, is then fed into the generative pooler and the contrastive pooler to get the final task-specific representations. This model corresponds to an early fusion of temporal information and does not add any new learnable layers. However, it makes the self-attention computation in the encoder much heavier due to the linearly increased number of tokens.

\vspace{-3mm}
\paragraph{Mean pooling.}
\label{sec:method_videococa_mean}
The frame-level representations $\mathbf{z}^{L} \in \mathbb{R}^{(B \times T) \times N \times d}$, are simply separately average pooled over the temporal dimension after the attentional poolers. It ignores the temporal information and is used as our adaptor baseline. Note that ~\cite{yu2022coca} also uses this method to compute zero-shot text-to-video retrieval metrics.

%\paragraph{Zero-shot transfer.}
%\label{sec:method_videococa_zeroshot_transfer}
\vspace{1mm}
\subsection{Finetuning VideoCoCa on Video-Text Data} \label{sec:method_tuning}
In addition to direct zero-shot transfer from image-text CoCa to video-text tasks, it is natural to consider further pushing the limit of VideoCoCa by continued pretraining on web-scale video-text paired data. We use VideoCC3M~\cite{nagrani2022learning} as our main source of video-text data which is detailed in Section~\ref{sec:experiments_setup}. We explore four different learning choices and conclude with the best setup with analysis.

\vspace{-3mm}
\paragraph{Finetuning (FT).} \label{sec:method_tuning_ft}
Under this setup, we unfreeze all parameters of the pretrained CoCa model during continued video-text pretraining, including the parameters of the encoder and the decoder, as well as the parameters of the generative pooler and the contrastive pooler. These parameters are finetuned together with the newly added learnable layers. Note that this could lead to less stable training~\cite{zhai2021lit}. We use it as our baseline tuning method.

\vspace{-3mm}
\paragraph{Frozen encoder-decoder tuning (Frozen).} \label{sec:method_tuning_ae}
We freeze the parameters of the encoder and the decoder, and only tune the parameters of the generative and the contrastive poolers. This allows us to re-use most parameters of the pretrained CoCa model.

\begin{figure}
    %\vspace{-0.5\baselineskip}
     \centering
     \begin{subfigure}[b]{0.49\linewidth}
         \centering
         \includegraphics[width=\textwidth]{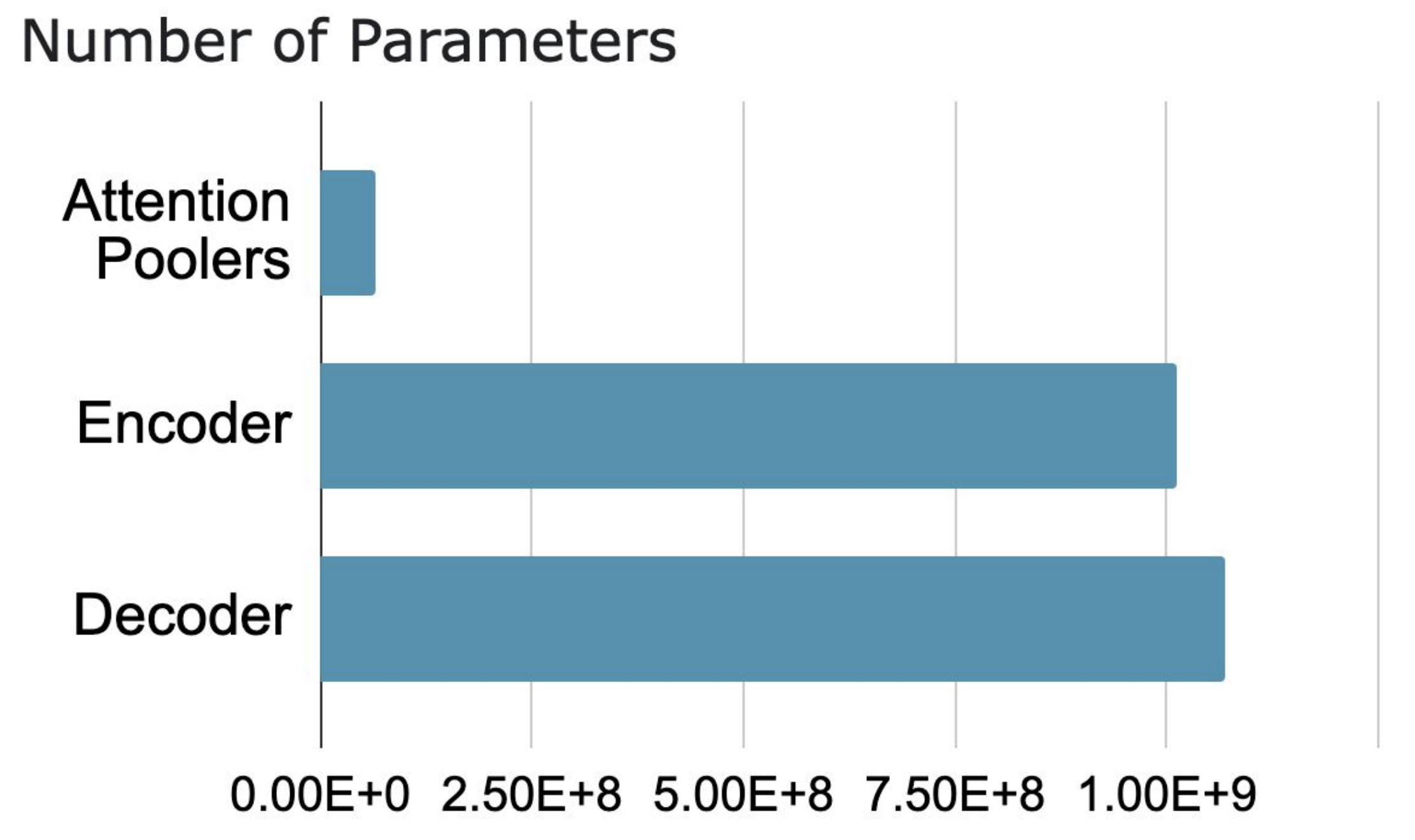}
         \caption{Total parameters of each component in CoCa.}
         \label{fig:params}
     \end{subfigure}
     \hfill
     \begin{subfigure}[b]{0.48\linewidth}
         \centering
         \includegraphics[width=\textwidth]{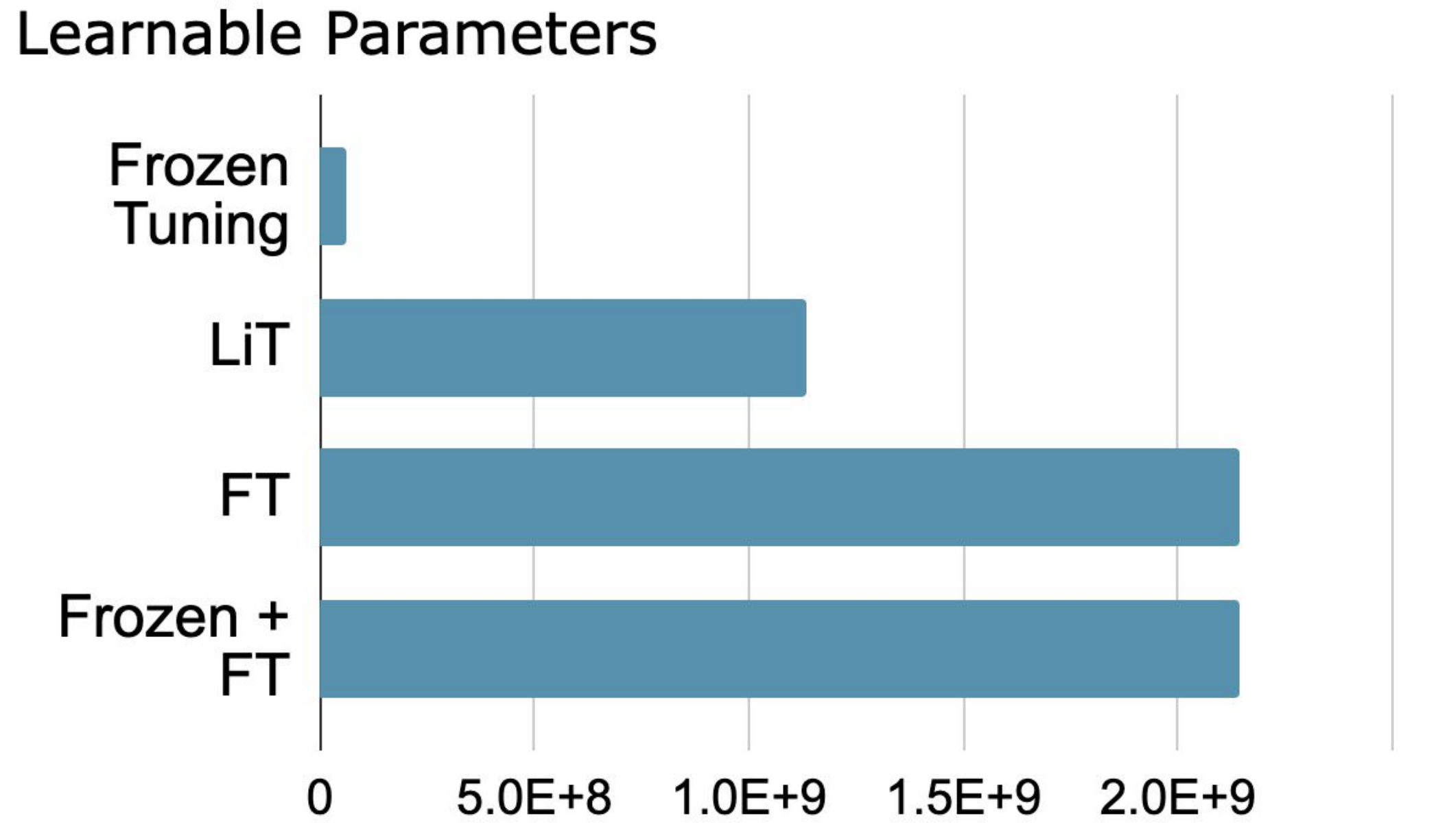}
         \caption{Learable parameters of different tuning methods.}
         \label{fig:tuning}
     \end{subfigure}
      \caption{An illustration of the parameters of each component (left), and learnable parameters of different tuning methods (right). LiT is adopted for VideoCoCa tuning.}
    \label{anaylysis}
    \vspace{-\baselineskip}
\end{figure}

\vspace{-3mm}
\paragraph{Frozen tuning then finetuning (Frozen + FT).}
\label{sec:method_tuning_ae_ft}
Frozen encoder-decoder tuning may converge very fast given the small amount of parameters of the pooler (See Figure~\ref{fig:params}). We explore to first conduct frozen feature tuning and then finetuning. In this way, the parameters of the pooler can be quickly trained, thus making the finetuning more stable. Note that it is a two-step tuning method. 

\vspace{-2mm}
\paragraph{Frozen encoder tuning (LiT).}
\label{sec:method_tuning_lit}
Similar to the best practice in \cite{zhai2021lit}, we only freeze the parameters of the pretrained CoCa image encoder, and tune the parameters of the poolers as well as the decoder. As the computation of image representations is much heavier than text representations, this not only allows us to precompute frame-level embeddings once to save TPU memory and computations for development, but also provides sufficient amount of learnable parameters for task adaptation (See Figure~\ref{fig:tuning}).

%\subsection{Scaling up batch size}
%\label{sec:method_batchsize}

\section{Experiments}
\label{sec:experiments}
%We first describe the details of our experimental setup in Section~\ref{sec:experiments_setup}. Next, we present the main results in Section~\ref{sec:experiments_sota} organized as zero-shot video classification, zero-shot text-video retrieval, zero-shot and finetuning results for video captioning, finetuning for video question answering, and model scalings. Finally, we present ablation experiments in Section~\ref{sec:experiments_ablation} including feature aggregation choices, tuning choices, and varying the number of frames. %Finally, We visualize the attention map of VideoCoCa and provide analysis in Sec. \ref{sec:experiments_visualizations}.

\subsection{Experimental setup}
\label{sec:experiments_setup}

\paragraph{Dataset for pretraining.} \label{sec:experiments_setup_upstream} In our experiments, we continue training the model on video-text datasets for one epoch to mitigate catastrophic forgetting. We report the performance of VideoCoCa models that are pretrained on the the following datasets:

\emph{HowTo100M}~\cite{miech2019howto100m} contains around 136 million video clips sourced from over 1.2M narrated videos associated with automatically transcribed speech.

\emph{VideoCC3M}~\cite{nagrani2022learning} is mined from the image captioning dataset Conceptual Captions 3M (CC3M)~\cite{sharma2018cc3m}. It includes around 8.7M video clip-text pairs and 900K unique captions. Unlike HowTo100M~\cite{miech2019howto100m}, which is restricted to instructional videos, VideoCC3M is created from a more diverse corpus of online videos and is thus more balanced. 
%We also experimented with mixing VideoCC3M and HowTo100M in each minibatch during pretraining, which is detailed in Appendix. 
% In the main text, we always use VideoCC3M as our pretraining dataset for the downstream tasks.

%\emph{HowTo100M}~\cite{miech2019howto100m} contains around 136 million video clips sourced from over 1.2M narrated videos associated with automatically transcribed speech. We obtain strong downstream results on nearby domain dataset Youcook2~\cite{zhou2017automatic}, but also observe a large domain gap on generic Youtube videos such as MSR-VTT~\cite{xu2016msrvtt}. See Appendix for details. 

%We also experimented with mixing VideoCC3M and HowTo100M in each minibatch in the pretraining, which is detailed in Appendix.  In the main text, we always use VideoCC3M as our pretraining dataset for the downstream tasks.

\vspace{-2mm}
\paragraph{Zero-shot video classification.}
\label{sec:experiments_setup_zeroshot_video_classification}
We use Kinetics\cite{kay_arxiv_2017}, UCF101~\cite{soomro2012ucf101}, HMDB51~\cite{kuehne2011hmdb}, and Charades~\cite{sigurdsson2016hollywood} to evaluate zero-shot transfer performance on video classification. We report results on Kinetics with 400, 600, and 700 classes, respectively. Following prior work~\cite{radford21clip,jia2021scaling}, we use aligned video/text embeddings to perform zero-shot video classification by matching videos with label names without
finetuning. We follow the same setup in~\cite{radford21clip} and apply the same set of prompts used for label class names. For Kinetics-400/600/700, we report Top-1 and Top-5 accuracy on the validation set. For UCF101 and HMDB51, we follow ~\cite{radford21clip,ni2022XCLIP} to evaluate on the three splits of the test data, and report the average Top-1 and Top-5 accuracy. For Charades, as it is a long-form video dataset with varied action classes over the duration (30 seconds in average), we follow~\cite{wang2021actionclip,bain2022cliphitchhikers} to report the multi-label classification results in mean average precision (mAP) on the test data. 

\vspace{-2mm}
\paragraph{Zero-shot text-video retrieval.}
\label{sec:experiments_setup_zeroshot_video_retrieval}
We report the zero-shot score of Recall at K (R@K) on MSR-VTT~\cite{xu2016msrvtt}, ActivityNet-Captions~\cite{krishna2017densecaptioning} and Youcook2~\cite{zhou2017automatic}. MSR-VTT contains 10K YouTube videos clips with 20 English descriptions each. We follow~\cite{portilloquintero2021straightforward,zeng2022socratic,yu2022coca} to report results on the available subset of the full test set. ActivityNet-Captions contains 20K YouTube videos with 100K captions. We evaluate paragraph-to-video retrieval~\cite{Luo2021CLIP4Clip, lei2022revealing} on the val1 split where the captions in the same video are concatenated to form a paragraph.  Youcook2 contains 2K cooking videos (14K video clips) and each video clip is annotated with one caption. We follow~\cite{miech2019howto100m,alayrac2022flamingo} to evaluate on the validation set. VATEX-v1.1~\cite{wang2019vatex} includes 26K/3K/6K videos for training/validation/test. There are 10 sentences in English and Chinese to describe each video. We follow ~\cite{chen2020finegrained,fang2021clip2video} to only use the English annotations for evaluation but on the entire validation set. 

\vspace{-2mm}
\paragraph{Video captioning.}
\label{sec:experiments_setup_ft_video_captioning}
We report both zero-shot and finetuning performance of BLEU-4~\cite{papineni2001bleu}, CIDEr~\cite{vedantam2014cider} and ROUGE~\cite{lin2004rouge} on MSR-VTT~\cite{xu2016msrvtt}, ActivityNet-Captions~\cite{krishna2017densecaptioning} and Youcook2~\cite{zhou2017automatic}. For MSR-VTT, we follow the standard splits in ~\cite{Luo2020UniVL,seo2022mvgpt} for a fair comparison. For ActivityNet-Captions, we concatenate captions for each segments to form a paragraph caption~\cite{lei2020mart,yamazaki2022vlcap} and use train/val1/val2 as training/validation/test splits.  
For Youcook2, we follow splits used in~\cite{Luo2020UniVL,xu2021vlm,alayrac2022flamingo} for finetuning and evaluation. For VATEX, we use train/validation/public test as the splits used in ~\cite{singh2020nitsvc,Zhang_2020_CVPR}.

\vspace{-2mm}
\paragraph{Finetuning for video question answering.}
\label{sec:experiments_setup_ft_video_qa}
We report the finetuning performance on the open-ended VideoQA datasets IVQA~\cite{yang2020just}, MSRVTT-QA~\cite{xu2017video},  MSVD-QA~\cite{xu2017video} and ActivityNet-QA~\cite{yu2019activityqa}. We use original splits for training, validation and test. We report top-1 test accuracy, which requires the model to learn a new task-specific attentional pooler on top of the decoder. Therefore, we only have finetuned results for the VideoQA task.

\vspace{-2mm}
\paragraph{Model variants.}
\label{sec:experiments_setup_model}
 For main results, we initialize the parameters of VideoCoCa from CoCa setup~\cite{yu2022coca} ("VideoCoCa" in short), where the model consists of 1B parameters in the image encoder and 1.1B parameters in the text decoder. We also explore Base and Large setup used in ~\cite{yu2022coca}, which are detailed in  Table~\ref{tab:model_scaling}. For ablation studies, we use the smaller model variant CoCa-Small in ~\cite{yu2022coca} where the decoder layers of CoCa-Base is reduced to 12. We fix 256 tokens as the generative embedding and 1 token as the contrastive embedding for all our experiments.

\paragraph{Training and inference.}
\label{sec:experiments_setup_training_inference}
We uniformly sample $8$ frames for each clip in all our experiments. Ablations with varying number of input frames are detailed in Appendix. We use the smaller model variant in ~\cite{yu2022coca} (\emph{i.e.} CoCa-Small) with the image resolution $224\times224$ and patch size $16\times16$ for our ablation study in Section~\ref{sec:experiments_ablation}. For the main results in Section~\ref{sec:experiments_sota}, we use CoCa with the image resolution $576\times576$ and patch size $18\times18$. For pretraining, we use a batch size of $128$ video-text pairs, and only apply center cropping to the training data. All models are trained on the combined contrastive and captioning objectives with the same optimization configuration as used in ~\cite{yu2022coca}, except with a lower learning rate of $1 \times 10^{-7}$ and weight decay of $1 \times 10^{-6}$. We continue pretraining VideoCoCa for one epoch. The training with precomputed frame token embeddings as input takes about $6$ hours on $128$ CloudTPUv4 chips. For finetuning, we use standard configurations and report the details in Appendix. During the evaluation, we always use a single view inference for simplicity. Specifically, we use the center cropping for each video frame, and no temporal croppings of a video is used.

\begin{table}[t]
\caption{Comparisons of 3 tasks zero-shot transfer performance with \textbf{different adaptors} on Kinetics-400 (video classification) and MSR-VTT (text-to-video retrieval, video captioning). All models are initialized from pretrained CoCa-Small and then finetuned on VideoCC3M. Without the need of learning new modules, attentional pooler consistently performs the best. \label{tab:feature_agg}}
\centering
\small
\setlength{\tabcolsep}{4pt}
\aboverulesep=0ex
\belowrulesep=0ex
%\vspace{2mm}
\resizebox{0.45\textwidth}{!}{%
%\renewcommand\arraystretch{1.05}
%\scriptsize{
\begin{tabular}{l|cc|cc|cc}
	\toprule 
 \multirow{2}{*}{\textbf{Adaptors}} & \multicolumn{2}{c|}{\textbf{VideoCls.}} & \multicolumn{2}{c|}{\textbf{VideoRet.}} & \multicolumn{2}{c}{\textbf{VideoCap.}} \\
 & \multirow{1}{*}{\footnotesize{Top-1}} & \multirow{1}{*}{\footnotesize{Top-5}} 
 & \multirow{1}{*}{\footnotesize{R@1}} & \multirow{1}{*}{\footnotesize{R@5}} 
  & \multirow{1}{*}{\footnotesize{BLEU-4}} 
 & \multirow{1}{*}{\footnotesize{CIDEr}} 
 \\
 \midrule
Mean Pooling & 40.3 & 69.3 & 24.5 & 45.3  & 15.5 & 13.4   \\ 
Factorized Enc. & 43.3 & 72.7 & 24.9 & 45.6  & 15.5 & 17.2   \\ 
%Unfactorized Enc. & 38.1 & 65.5 & 22.6 & 43.8 & 55.5 & 14.7 & 12.1 & 21.6  \\ 
Joint Space-Time & 38.1 & 65.5 & 22.6 & 43.8  & 14.7 & 12.1  \\ 
Attentional Pooler & \textbf{45.6} & \textbf{73.4} & \textbf{26.4} & \textbf{46.8}  & \textbf{16.8} & \textbf{19.9} \\ 
 \bottomrule
\end{tabular}
}
%}
\vspace{-1\baselineskip}
\end{table}

%\vspace{\baselineskip}
\subsection{Ablation study}
\label{sec:experiments_ablation}
We conduct ablation studies on the Kinetics-400, MSR-VTT, ActivityNet-Captions and Youcook2. We report zero-shot video classification accuracies, text-to-video retrieval R@K and video captioning results. 

\vspace{-2mm}
\paragraph{Adaptor choices.}
\label{sec:experiments_feature_agg_ablation}
Table \ref{tab:feature_agg} shows the comparison of different adaptation methods. In all cases, the model is initialized from pretrained CoCa-Small and then continued pretraining on VideoCC3M using "Finetuning" method in Section~\ref{sec:method_tuning}. Compared to the "Mean Pooling" baseline, which ignores temporal information, "Factorized Encoder" clearly outperforms on video classification ($+3\%$ and $+3.4\%$ on top-1 and top-5 accuracy, respectively) and video captioning ($+3.8$ CIDEr score), indicating that a late fusion of temporal information is useful to learn high-quality video representation better than the image-text approach. We observe that "Attentional Pooler", which also corresponds to the late fusion but does not introduce any new video-specific parameters, outperforms all other methods on three video-text transfer tasks ($+5.3\%$ top-1 on video classification, $+1.9$ R@1 on text-to-video retrieval, $+6.5$ CIDEr on video captioning). "Joint Space-Time Encoder", which corrsponds to an early fusion of temporal information, is the worst performing method of the four, and we think it is because early reshaping of the spatial-temporal features causes the input pattern to change, making the pretrained image-text model harder to adapt. Given the strong zero-shot transfer results from "Attentional Pooler" and its minimalist design, we use it in subsequent experiments.

\begin{table}[t]
\caption{Comparisons of 3 tasks zero-shot transfer performance with \textbf{different tuning choices} on Kinetics-400 (video classification) and MSR-VTT (text-to-video retrieval, video captioning). All models are initialized from a pretrained CoCa-Small. Frozen image encoder tuning (LiT) is adopted as it allows faster and lightweight task adaptation. \label{tab:tuning_choices}}
\centering
\small
\setlength{\tabcolsep}{4pt}
\aboverulesep=0ex
\belowrulesep=0ex
%\resizebox{0.49\textwidth}{!}{%
%\renewcommand\arraystretch{1.05}
%\scriptsize{
\resizebox{0.42\textwidth}{!}{%
\begin{tabular}{l|cc|cc|cc}
	\toprule 
 \multirow{2}{*}{\textbf{Tuning choices}} & \multicolumn{2}{c|}{\textbf{VideoCls.}} & \multicolumn{2}{c|}{\textbf{VideoRet.}} & \multicolumn{2}{c}{\textbf{VideoCap.}} \\
 & \multirow{1}{*}{\footnotesize{Top-1}} & \multirow{1}{*}{\footnotesize{Top-5}} 
 & \multirow{1}{*}{\footnotesize{R@1}} & \multirow{1}{*}{\footnotesize{R@5}} 
 & \multirow{1}{*}{\footnotesize{BLEU-4}} 
 & \multirow{1}{*}{\footnotesize{CIDEr}}  
 \\
 \midrule
FT & 45.6 & 73.4 & 26.4 & 46.8  & 16.8 & 19.9   \\ 
Frozen & 45.4 & 73.2 & 27.3 & 47.0  & 16.2 & 19.7   \\ 
Frozen + FT & \textbf{45.8} & 73.5 & 28.4 & 49.3  & 16.8 & 20.5   \\ 
LiT & 45.7 & \textbf{73.9} & \textbf{28.8} & \textbf{49.3}  & \textbf{17.0} & \textbf{20.6}  \\ 
 \bottomrule
\end{tabular}
%}
}
\vspace{-1\baselineskip}
\end{table}

\vspace{-2mm}
\paragraph{Tuning choices.}
\label{sec:experiments_tuning_ablation}
Table \ref{tab:tuning_choices} shows the comparison on zero-shot video-text tasks transfer with different tuning choices. VideoCoCa is initialized from CoCa-Small and continue pretrained on VideoCC3M. Compared to "Finetuning" (FT) baseline, we surprisingly find that "Frozen Encoder-Decoder  Tuning"  (Frozen) alone already achieves comparable results to the baseline on all tasks. This allows us to freeze most of the parameters of CoCa and only tune a few parameters in the attentional pooler. "Frozen Tuning then Finetuning" (Frozen + FT), is able to further improve the performance on top of frozen tuning ($+2.0$ on R@1), however, is inconvenient in practice as it needs to first conduct frozen tuning then finetuning. Frozen Encoder Tuning (LiT), which freezes the parameters of image encoder and only tunes the parameters of text decoder and attentional poolers, achieves the best performance on text-to-video retrieval ($+2.4$ on R@1) and video captioning ($+0.7$ on CIDEr) among four tuning methods, indicating that a strong frozen image encoder can be enough for video-text tasks adaptation. Another advantage is that it allows us to precompute frame-level embeddings, and thus saves most of the TPU memory in the actual training. Therefore, we adopt LiT for subsequent experiments.

\vspace{-2mm}
\paragraph{Pretraining with mixed VideoCC3M and HowTo100M.}
\label{sec: mixed_ht_hc}
We experimented with mixing VideoCC3M and HowTo100M in each minibatch during pretraining. The model is initialized from the pretrained CoCa-Small and then continued pretraining on the mixed datasets. Table~\ref{tab:mix_ablations} shows results on zero-shot text-to-video retrieval in R@10. We find that increasing the mixing ratio of HowTo100M in each minibatch gives monotonically increased results on nearby domain dataset YouCook2 but decreased results on generic Youtube videos MSR-VTT and ActivityNet Captions. For example, comparing to using 100\% VideoCC3M for pretraining, using 100\% HowTo100M gives $+1.7\%$ improvement on Youcook2, but $-2.5\%$ on MSR-VTT and $-6.0\%$ on ActivityNet Captions. We observed that there is no optimal mixing ratio that provides the best results on all three datasets. Therefore, we always use VideoCC3M as our pretraining dataset.

\begin{table}[t]
    \centering
    \setlength{\tabcolsep}{3pt}
    \caption{
    Results on zero-shot text-to-video retrieval with \textbf{different mixing ratios} between VideoCC3M and HowTo100M in each minibatch of pretraining. All results are reported in R@10 on MSR-VTT Full, ActivityNet-Captions and Youcook2. All VideoCoCa models are initialized from a pretrained CoCa-Small model then trained with LiT.}
    
    \resizebox{\columnwidth}{!}{%
   % \footnotesize{
    \begin{tabular}{ccccc}
    \toprule
    \multicolumn{2}{c}{Mixing Ratio [\%]} & MSR-VTT Full & ActivityNet-Captions & YouCook2  \\
    VideoCC3M & HT100M & & & \\   \midrule 
    100 & 0   & \textbf{59.5} & \textbf{66.8} & 29.3 \\
    70 & 30    & 59.0 & 66.1 & 29.6 \\
    50 & 50    & 58.1 & 65.6 & 30.0 \\
    30 & 70    & 57.6 & 63.4 & 30.5 \\
    0 & 100    & 57.0 & 60.8 & \textbf{31.0} \\
    \bottomrule
    \label{tab:mix_ablations}
    \end{tabular}
    %}
    }
    \vspace{-2\baselineskip}
\end{table}

\begin{table*}[t]
    \vspace{-1.0\baselineskip}
	\caption{Comparisons to state-of-the-art results on zero-shot video classification. Results are reported in top accuracy [\%] or mean average precision (mAP). \emph{Avg.} denotes averaged Top-1/5 accuracy. "CoCa" refers to CoCa image-text model that is directly applied to zero-shot video classification using the frame-level average pooling. We use the notation \textit{``w.o. videos"} to refer to the VideoCoCa model that is initialized from the pretrained CoCa model without continued pretraining. In line with~\cite{ni2022XCLIP}, we use single view evaluation for simplicity.}
	\vspace{-0.5\baselineskip}
	\begin{subtable}[t]{.31\linewidth}
		\centering
		\caption{Kinetics 400}
    	\setlength{\tabcolsep}{4pt} %
		\renewcommand*{\arraystretch}{1.10}  %
		\vspace{-0.3\baselineskip}
		\scriptsize{
			\begin{tabular}{lcc}
				\toprule
				Method 	& Top 1 & Top 5 \\
				\midrule
				REST~\cite{bulat2022rest} & 63.9 & 81.0 \\
				\midrule                             
				CoCa &  66.4 &  87.1  \\  %
				\textbf{VideoCoCa} (\textit{w.o. videos}) & \textbf{72.0} & 90.3 \\ %
				\textbf{VideoCoCa} 	& \textbf{72.0} & \textbf{90.5}  \\ %
				\bottomrule
			\end{tabular}
		}
		\label{tab:sota_kinetics400}
	  	\vspace{0.96\baselineskip} %
  		\centering
  		\caption{HMDB 51}
  		\vspace{-0.2\baselineskip}
  		\scriptsize{
  			\begin{tabular}{lcc}
  				\toprule Method
  				& Top 1 & Top 5 \\ 
  				\midrule
  				ER-ZSAR~\cite{chen2021elaborative} 	&  35.3   &  --     \\
  				ActionCLIP~\cite{wang2021actionclip} & 40.8 & -- \\  %
  				X-CLIP~\cite{ni2022XCLIP}	&  44.6 &  --     \\
  				X-Florence~\cite{ni2022XCLIP} & 48.4 & -- \\  %
  				REST~\cite{bulat2022rest} & 49.7 & -- \\  % 
  				MOV-L~\cite{qian2022multimodal} & 57.8 & -- \\
  				\midrule
  				CoCa &  57.2 &  82.6  \\  %
				\textbf{VideoCoCa} (\textit{w.o. videos}) & 57.4 & 82.7 \\ %
				\textbf{VideoCoCa} 	& \textbf{58.7} & \textbf{84.5}  \\ %
  				\bottomrule
  			\end{tabular}
  			\label{tab:sota_hmdb51}
  		}	
	\end{subtable}
  	%\hfill
  	\begin{subtable}[t]{.31\linewidth}
		\centering
  		\caption{Kinetics 600}
  		\setlength{\tabcolsep}{4pt} %
		\vspace{-0.3\baselineskip}
  		\scriptsize{
	  		\begin{tabular}{lcc}
	  			\toprule
	  			Method 	& Top 1  & Top 5  \\ %
	  			\midrule
	  			ER-ZSAR~\cite{chen2021elaborative} & 42.1  & 73.1 \\ %
				X-CLIP~\cite{ni2022XCLIP} & 65.2  & 86.1 \\ %
				X-Florence~\cite{ni2022XCLIP} & 68.8  & 88.4 \\ %
	  			\midrule
  				CoCa & 65.1 & 87.1  \\
	  			\textbf{VideoCoCa} (\textit{w.o. videos}) & 70.0 & \textbf{89.0} \\ %
				\textbf{VideoCoCa} 	& \textbf{70.1} & 88.9  \\ %
	  			\bottomrule
	  		\end{tabular}
	  		\label{tab:sota_kinetics600}
  		}
  		\vspace{0.3\baselineskip} %
		\centering
		\caption{UCF 101}
		\vspace{-0.2\baselineskip}
			\scriptsize{
  			\begin{tabular}{lcc}
  				\toprule Method
  				& Top 1 & Top 5 \\ 
  				\midrule
  				ER-ZSAR~\cite{chen2021elaborative} 	&  51.8  &  --     \\
  				ActionCLIP~\cite{wang2021actionclip} & 58.3 & -- \\  %
  				
  				REST~\cite{bulat2022rest} & 69.1 & -- \\  %
  				X-CLIP~\cite{ni2022XCLIP}	&  72.0 &  --     \\
  				X-Florence~\cite{ni2022XCLIP} & 73.2 & -- \\  %
  				MOV-L~\cite{qian2022multimodal} & 80.9 & -- \\
  				\midrule
  				CoCa &  85.1 &  97.5  \\  %
				\textbf{VideoCoCa} (\textit{w.o. videos}) & \textbf{86.6} & \textbf{98.5} \\ %
				\textbf{VideoCoCa} 	& \textbf{86.6} & 98.4  \\ %
  				\bottomrule
  			\end{tabular}
  			\label{tab:sota_ucf101}
  		}	
  	\end{subtable}
  	\hfill
  	\begin{subtable}[t]{.35\linewidth}
  		\setlength{\tabcolsep}{4pt} %
  		\centering
  		\caption{Kinetics 700}
  		\vspace{-0.3\baselineskip}
  		\setlength{\tabcolsep}{6pt} %
  		\scriptsize{
  			\begin{tabular}{lccc}
  				\toprule 
  				Method & Top 1 & Top 5 & Avg. \\ 
  				\midrule
  				NFNet-F6~\cite{alayrac2022flamingo} & -- & -- & 62.9 \\  %
  				CLIP-L~\cite{alayrac2022flamingo} & -- & -- & 69.6 \\
  				\midrule
	  			CoCa & 60.2 & 82.2 & 71.2  \\
	  			\textbf{VideoCoCa} (\textit{w.o. videos}) & 62.2 & 83.7 & 73.0 \\ %
				\textbf{VideoCoCa} 	& \textbf{62.5} & \textbf{83.8} & \textbf{73.2} \\ %
  				\bottomrule
  			\end{tabular}
  			\label{tab:sota_kinetics700}
  		}
  		\vspace{0.8\baselineskip} %
  		\centering
  		\caption{Charades}
  		\vspace{-0.2\baselineskip}
		\scriptsize{
  			\begin{tabular}{lcc}
  				\toprule 
  				Method & mAP\\ 
  				\midrule
  				CLIP-Hitchhiker~\cite{bain2022cliphitchhikers} & 21.1 \\
  				\midrule
	  			CoCa & 23.1  \\
	  			\textbf{VideoCoCa} (\textit{w.o. videos}) & 25.2\\ %
				\textbf{VideoCoCa} 	& \textbf{25.8}  \\ %
  				\bottomrule
  			\end{tabular}
  			\label{tab:sota_charades}
  		}
	\end{subtable}
	\label{tab:sota_cls}
	\vspace{-0.5\baselineskip}
\end{table*}

\begin{table*}[t]
\vspace{1.\baselineskip}
\caption{Model scaling results. We report results on zero-shot video classification (top-1 accuracy), zero-shot text-to-video retrieval (R@1), zero-shot video captioning (CIDEr) and finetuning for video question answering (top-1 accuracy) across three model scales. VideoCoCa consistently outperforms CoCa baseline (mean pooling) with the same number of parameters.\label{tab:model_scaling}}
\centering
\small
\setlength{\tabcolsep}{2pt}
\aboverulesep=0ex
\belowrulesep=0ex
\vspace{-0.5\baselineskip}
\resizebox{0.99\textwidth}{!}{%
\begin{tabular}{l|c|c|ccccc|ccc|ccc|ccc}
	\toprule 
 \multirow{2}{*}{\textbf{Model}} & \multirow{2}{*}{\textbf{Params}} & \multirow{2}{*}{\textbf{TFLOPs}} & \multicolumn{5}{c|}{\textbf{VideoCls.}} & \multicolumn{3}{c|}{\textbf{VideoRet.}} & \multicolumn{3}{c|}{\textbf{VideoCap.}} & \multicolumn{3}{c}{\textbf{VideoQA}} \\ & &
 & \multirow{1}{*}{\footnotesize{K400}} & \multirow{1}{*}{\footnotesize{K600}} 
 & \multirow{1}{*}{\footnotesize{K700}} & \multirow{1}{*}{\footnotesize{HMDB51}} 
 & \multirow{1}{*}{\footnotesize{UCF101}} &  \multirow{1}{*}{\footnotesize{MSRVTT}}
 & \multirow{1}{*}{\footnotesize{ActivityNet}} 
 & \multirow{1}{*}{\footnotesize{Youcook2}} & \multirow{1}{*}{\footnotesize{MSRVTT}}
  & \multirow{1}{*}{\footnotesize{ActivityNet}} 
  & \multirow{1}{*}{\footnotesize{Youcook2}}
 & \multirow{1}{*}{\footnotesize{IVQA}} & \multirow{1}{*}{\footnotesize{MSRVTT-QA}} 
 & \multirow{1}{*}{\footnotesize{MSVD-QA}} \\
 \midrule
CoCa-B & 383M & 2.01 & 57.5 & 56.5 & 47.6 & 51.1 & 78.3 & 27.5 & 21.7 & 11.2 & 12.8 & 8.5 & 7.0 & 28.7 & 41.2 & 52.2  \\
\textbf{VideoCoCa-B} & 383M & 2.22 & \textbf{61.2} & \textbf{59.6} & \textbf{51.0} & \textbf{53.7} & \textbf{80.4} & \textbf{31.2} & \textbf{29.6} & \textbf{16.5} & \textbf{20.6} & \textbf{14.1} & \textbf{14.4} & \textbf{29.4} & \textbf{42.6} & \textbf{53.6}  \\ \midrule
CoCa-L & 787M & 6.89 & 62.9 & 60.7 & 50.5 & 52.2 & 84.6 & 29.0 & 24.2 & 13.8 & 16.5 & 9.1 & 15.7 & 31.2 & 43.3 & 55.0  \\
\textbf{VideoCoCa-L} & 787M & 7.39 & \textbf{68.4} & 
\textbf{67.4} & \textbf{58.9} & \textbf{56.9} & \textbf{85.3} & \textbf{33.9} & \textbf{31.5} & \textbf{18.9} & \textbf{24.3} & \textbf{17.4} & \textbf{20.7} & \textbf{32.6} & \textbf{44.3} & \textbf{56.2}  \\
\midrule
CoCa & 2.1B & 28.74 & 66.4 & 65.1 & 60.2 & 57.2 & 85.1 & 30.0 & 28.5 & 16.8 & 15.9 & 11.2 & 18.2 & 33.6 & 45.9 & 55.4  \\
\textbf{VideoCoCa} & 2.1B & 29.82 & \textbf{72.0} & \textbf{70.1} & \textbf{62.5} & \textbf{58.7} & \textbf{86.6} & \textbf{34.3} & \textbf{34.5} & \textbf{20.3} & \textbf{27.1} & \textbf{19.3} & \textbf{34.3} & \textbf{39.0} & \textbf{46.3} & \textbf{56.9}  \\

 \bottomrule
\end{tabular}
}
\end{table*}

\begin{table*}[t]
    %\vspace{1\baselineskip}
	\caption{Comparison to state-of-the-art results on zero-shot text-video retrieval. All results are reported on R@1 / R@5 / R@10. ``\textit{w.o. videos}" refers to the VideoCoCa model that is initialized from the pretrained CoCa model without continued pretraining. For Youcook2 results, see Table \ref{tab:mix_ablations} for the analysis.}
	\vspace{-1.2\baselineskip}
	\begin{tabular}{cc}
	\begin{subtable}[t]{0.49\linewidth}
		\centering
		\caption{MSR-VTT Full}
    	\setlength{\tabcolsep}{2pt} %
		\renewcommand*{\arraystretch}{1.10}  %
		\vspace{-0.3\baselineskip}
		\scriptsize{
	
			\begin{tabular}{l|ccc|ccc}
				\toprule 
 \multirow{2}{*}{Method} & \multicolumn{3}{c|}{Text-to-Video} & \multicolumn{3}{c}{Video-to-Text} \\
 & \multirow{1}{*}{\footnotesize{R@1}} & \multirow{1}{*}{\footnotesize{R@5}} 
 & \multirow{1}{*}{\footnotesize{R@10}} & \multirow{1}{*}{\footnotesize{R@1}} 
 & \multirow{1}{*}{\footnotesize{R@5}}  & \multirow{1}{*}{\footnotesize{R@10}}  
 \\
 \midrule
CLIP~\cite{portilloquintero2021straightforward} & 23.3 & 44.2 & 53.6 & 43.3 & 73.3 & 81.8  \\ 
Socratic Models~\cite{zeng2022socratic} & -- & -- & -- & 46.9 & 73.5 & 81.3  \\ 
CoCa~\cite{yu2022coca} & 30.0 & 52.4 & 61.6 & 49.9 & 73.4 & 81.4  \\ \midrule
\textbf{VideoCoCa} (\textit{w.o. videos}) & 32.9 & 56.0 & 65.0 & 59.7 & 81.8 & 89.3  \\ 
\textbf{VideoCoCa} & \textbf{34.3} & \textbf{57.8} & \textbf{67.0} & \textbf{64.7} & \textbf{85.2} & \textbf{91.4}  \\ 
 \bottomrule
\end{tabular} \label{tab:sota_msrvtt_retrieval}
}
	\end{subtable}
	
  %	\hfill

  	\begin{subtable}[t]{0.49\linewidth}
  	\vspace{-0.35\baselineskip}
		\centering
  		\caption{ActivityNet Captions}
  		\setlength{\tabcolsep}{2pt} %
		\vspace{0.1\baselineskip}
  		\scriptsize{
	\begin{tabular}{l|ccc|ccc}
		\toprule 
 \multirow{2}{*}{Method} & \multicolumn{3}{c}{Text-to-Video} & \multicolumn{3}{c}{Video-to-Text} \\
 & \multirow{1}{*}{\footnotesize{R@1}} & \multirow{1}{*}{\footnotesize{R@5}} 
 & \multirow{1}{*}{\footnotesize{R@10}} & \multirow{1}{*}{\footnotesize{R@1}} 
 & \multirow{1}{*}{\footnotesize{R@5}}  & \multirow{1}{*}{\footnotesize{R@10}}  
 \\
 \midrule
SINGULARITY~\cite{lei2022revealing}  & 30.6 & 55.6 & 66.9 & -- & -- & -- \\ 
InternVideo~\cite{wang2022internvideo} & 30.7 & -- & -- & 31.4 & -- & --  \\
\midrule
CoCa & 28.5 & 58.4 & 70.5 & 21.9 & 49.5 & 64.1 \\
\textbf{VideoCoCa} (\textit{w.o. videos}) & 32.2 & 61.0 & 74.6 & 26.8 & 54.7 & 70.0  \\ 
\textbf{VideoCoCa} & \textbf{34.5} & \textbf{63.2} & \textbf{76.6} & \textbf{33.0} & \textbf{61.6} & \textbf{75.3}   \\ 

 \bottomrule
\end{tabular}	\label{tab:sota_activitynet_retrieval}
  		}
  	\end{subtable}
  	\end{tabular}
  %	\hfill
  \begin{tabular}{cc}

	  	\begin{subtable}[t]{0.48\linewidth}
  		\setlength{\tabcolsep}{2pt} %
  		\centering
  		\caption{VATEX}
  		%\vspace{-0.3\baselineskip}
  		%\setlength{\tabcolsep}{4pt} %
  		\scriptsize{
  			\begin{tabular}{l|ccc|ccc}
  				\toprule 
 \multirow{2}{*}{Method} & \multicolumn{3}{c}{Text-to-Video} & \multicolumn{3}{c}{Video-to-Text} \\
 & \multirow{1}{*}{\footnotesize{R@1}} & \multirow{1}{*}{\footnotesize{R@5}} 
 & \multirow{1}{*}{\footnotesize{R@10}} & \multirow{1}{*}{\footnotesize{R@1}} 
 & \multirow{1}{*}{\footnotesize{R@5}}  & \multirow{1}{*}{\footnotesize{R@10}}  
 \\ \midrule
InternVideo~\cite{wang2022internvideo} & 49.5 & -- & -- & 69.5 & -- & -- \\ \midrule
CoCa & 47.2 & 79.0 & 86.9 & 60.9 & 88.1 & 82.4  \\ 
\textbf{VideoCoCa} (\textit{w.o. videos}) & 50.7 & 81.3 & 88.5 & 64.1 & 90.2 & 95.6 \\ 
\textbf{VideoCoCa} & \textbf{53.2} & \textbf{83.3} & \textbf{90.1} &  \textbf{73.6} & \textbf{93.2} & \textbf{97.2}  \\ 
 \bottomrule
\end{tabular} \label{tab:sota_vatex_retrieval}
  		}
	\end{subtable}
	
  	 %	\hfill

  	\begin{subtable}[t]{0.48\linewidth}
  	\vspace{-0.5\baselineskip}
  		\setlength{\tabcolsep}{2pt} %
  		\centering
  		\caption{Youcook2}
  		\vspace{-0.3\baselineskip}
  		\scriptsize{
  			\begin{tabular}{l|ccc}
  				\toprule 
 \multirow{2}{*}{Method} & \multicolumn{3}{c}{Text-to-Video}  \\
 & \multirow{1}{*}{\footnotesize{R@1}} & \multirow{1}{*}{\footnotesize{R@5}} 
 & \multirow{1}{*}{\footnotesize{R@10}} \\
 \midrule
FitCLIP~\cite{castro2022fitclip} & 5.8 & 15.5 & 22.1  \\ 
TACo~\cite{yang2021taco} & 19.9 & 43.2 & 55.7  \\ 
VideoCLIP~\cite{xu2021videoclip} & \textbf{22.7} & \textbf{50.4} & \textbf{63.1} \\ \midrule
CoCa & 16.8 & 35.8 & 46.4  \\ 
\textbf{VideoCoCa} (\textit{w.o. videos}) & 21.7 & 43.9 & 55.2 \\ 
\textbf{VideoCoCa} & 20.3 & 43.0 & 53.3 \\ 
%VideoCoCa (HT100M) & 22.6 & 47.0 & 59.3 \\ 
 \bottomrule
\end{tabular} \label{tab:sota_youcook2_retrieval}
  		}
	\end{subtable}
	  \end{tabular}

	\label{tab:sota_ret}
	\vspace{-1.\baselineskip}
\end{table*}

%VideoCap results
\begin{table*}[t]
    \vspace{1\baselineskip}
	\caption{Comparison to state-of-the-art zero-shot and finetuned results on video captioning.  All results are reported on BLEU-4 (B-4), CIDEr (C) and ROUGE (R).}
	\vspace{-1.2\baselineskip}
	\begin{tabular}{cc}
	\begin{subtable}[t]{0.24\linewidth}
		\centering
		\caption{MSR-VTT}
    	\setlength{\tabcolsep}{3pt} %
		%\renewcommand*{\arraystretch}{1.10}  %
		%\vspace{-0.3\baselineskip}
		\scriptsize{
			\begin{tabular}{l|ccc}
\toprule 
  			Method & B-4 & C & R\\ 
 
 \midrule
\multicolumn{4}{l}{\textit{zero-shot results}}        \\
AVR ~\cite{nagrani2022learning} & 13.2 & 8.2 & --  \\ 
\textbf{VideoCoCa} &  \textbf{19.7} & \textbf{27.1} & \textbf{35.4}  \\ \midrule
\multicolumn{4}{l}{\textit{finetuning results}}        \\
UniVL~\cite{Luo2020UniVL} & 41.8 & 50.0 & 60.8 \\
MV-GPT~\cite{seo2022mvgpt} & 48.9 & 60.0 & 64.0 \\
GIT2~\cite{wang2022git} & \textbf{54.8} & \textbf{75.9} & \textbf{68.2} \\
\textbf{VideoCoCa} & 53.8 & 73.2 & 68.0  \\ 
 \bottomrule
\end{tabular} \label{tab:sota_msrvtt_captioning}
}
		
	\end{subtable}
  %\hfill
  	%\vspace{-0.5\baselineskip}
  	
  	\begin{subtable}[t]{0.24\linewidth}
		\centering
  		\caption{ActivityNet Captions}
  		\setlength{\tabcolsep}{3pt} %
		%\vspace{0.1\baselineskip}
		\scriptsize{
			\begin{tabular}{l|ccc}
\toprule 
  			Method & B-4 & C & R\\ 
 \midrule
\multicolumn{4}{l}{\textit{zero-shot results}}  \\ 
CoCa & 6.9 & 11.2 & 19.8 \\
\textbf{VideoCoCa} & \textbf{8.5} & \textbf{19.3} & \textbf{22.5}   \\ \midrule
\multicolumn{4}{l}{\textit{finetuning results}}        \\
MART~\cite{lei2020mart} & 9.8 & 22.2 & 30.9 \\
VLCap~\cite{yamazaki2022vlcap} & 13.4 & 30.3 & \textbf{36.0} \\
\textbf{VideoCoCa} & \textbf{14.7} & \textbf{39.3} & 35.0 \\
 \bottomrule
\end{tabular} \label{tab:sota_activitynet_captioning}
}
  	\end{subtable}

	%\end{tabular}

 	%\begin{tabular}{cc}
 	%\vspace{-0.5\baselineskip}
  	\begin{subtable}[t]{0.25\linewidth}
  		\setlength{\tabcolsep}{3pt} %
  		\centering
  		\caption{VATEX}
		\scriptsize{
			\begin{tabular}{l|ccc}
            \toprule 
  			Method & B-4 & C & R\\ 
            \midrule
\multicolumn{4}{l}{\textit{zero-shot results}}        \\
CoCa & 6.2 & 12.4 & 21.0  \\ 
\textbf{VideoCoCa} &  \textbf{9.5} & \textbf{22.8} & \textbf{29.5}  \\ \midrule
\multicolumn{4}{l}{\textit{finetuning results}}        \\
NITS-VC~\cite{singh2020nitsvc} & 20.0 & 24.0 & 42.0 \\
ORG-TRL~\cite{Zhang_2020_CVPR} & 32.1 & 49.7 & 48.9 \\
\textbf{VideoCoCa} & \textbf{39.7} & \textbf{77.8} & \textbf{54.5}  \\ 
 \bottomrule
\end{tabular} \label{tab:sota_vatex_captioning}
}
	\end{subtable}

  	\begin{subtable}[t]{0.25\linewidth}
  	 %\vspace{-1\baselineskip}
  		\setlength{\tabcolsep}{2pt} %
  		\centering
  		\caption{Youcook2}
  		%\vspace{-0.3\baselineskip}
  		%\setlength{\tabcolsep}{4pt} %
		\scriptsize{
			\begin{tabular}{l|ccc}
\toprule 
  			Method & B-4 & C & R\\ 
 
 \midrule
\multicolumn{4}{l}{\textit{zero-shot results}}        \\
Flamingo-80B ~\cite{alayrac2022flamingo} & -- & 60.1 & --  \\ 
\textbf{VideoCoCa} &  7.7 & 34.3 & 16.5  \\ \midrule
\multicolumn{4}{l}{\textit{finetuning results}}        \\
UniVL~\cite{Luo2020UniVL} & 11.2 & 127.0 & 40.1 \\
VLM~\cite{xu2021vlm} & 12.3 & \textbf{138.7} & \textbf{41.5} \\
OmniVL~\cite{wang2022omnivl} & 8.7 & 116.0 & 36.1 \\
GIT2~\cite{wang2022git} & 9.4 & 131.2 & 37.5 \\
\textbf{VideoCoCa} & \textbf{14.2} & 128.0 & 37.7  \\ 
 \bottomrule
\end{tabular} \label{tab:sota_youcook2_captioning}
}
	\end{subtable}
	\end{tabular}

	\label{tab:sota_cap}
%\vspace{-1\baselineskip}
\end{table*}

\begin{table}[t]
\caption{Comparison to state-of-the-art results on finetuning for video question answering. All results are in top-1 accuracy [\%].}
	\vspace{-\baselineskip}
\centering
\small
\setlength{\tabcolsep}{3pt}
\aboverulesep=0ex
\belowrulesep=0ex
\vspace{2mm}
\resizebox{0.49\textwidth}{!}{%
\renewcommand\arraystretch{1.05}

\begin{tabular}{l|cccc}
\toprule
Method & IVQA & MSRVTT-QA & MSVD-QA & ActivityNet-QA\\
 \midrule
JustAsk~\cite{yang2020just} & 35.4 & 41.8 & 47.5 & 38.9 \\
SINGULARITY~\cite{lei2022revealing} & -- & 43.5 & -- & 44.1 \\
VIOLET~\cite{fu2021violet} & -- & 43.9 & 47.9 & -- \\
All-in-one~\cite{wang2022allinone} & -- & 46.8 & 48.3 & -- \\
FrozenBiLM~\cite{yang2022zeroshot} & \textbf{39.6} & \textbf{47.0} & 54.8 & 43.2 \\
GiT2~\cite{wang2022git} & -- & 45.6 & \textbf{58.2} & --\\
\midrule
\textbf{VideoCoCa} & 39.0 & 46.3 & 56.9 & \textbf{56.1} \\
\bottomrule
\end{tabular}
}
\label{tab:sota_qa}
\vspace{-0.8\baselineskip}
\end{table}

\vspace{-1\baselineskip}
\paragraph{Scaling up the model.} \label{sec:experiments_ablation_scaling_models}
To make the notation consistent with ~\cite{yu2022coca}, we use VideoCoCa, VideoCoCa-L, and VideoCoCa-B to refer to model parameters initialized from CoCa, CoCa-Large, and CoCa-Base, respectively. Table ~\ref{tab:model_scaling} shows our model scaling results on four video-text tasks. All models are trained with a spatial resolution of $576\times576$ and the total FLOPs (TFLOPs) are reported. With the same number of parameters, VideoCoCa consistently achieves higher performance than the CoCa model at every model scale. Furthermore, note how our Large model (transformer depth
of 24 layers) outperforms CoCa (40 layers), while using 3.9$\times$ less TFLOPs. Similarly, our Base model outperforms CoCa-L on most datasets. This result indicates that we can achieve greater video-text tasks improvements by adapting attentional poolers for videos with continued training. When using our largest 2.1B model, our VideoCoCa substantially advances the best results obtained from CoCa with less than 4\% extra TFLOPs compute.

%\vspace{\baselineskip}
\subsection{Main Results}
\label{sec:experiments_sota}
We compare VideoCoCa to the state-of-the-arts on fourteen different datasets and four tasks, including zero-shot video classification, zero-shot text-video retrieval, zero-shot and finetuning for video captioning, and finetuning for video question answering. We evaluate models with single view per video clip, following~\cite{ni2022XCLIP} (detailed in Section~\ref{sec:experiments_setup_training_inference}).

\vspace{-1\baselineskip}
\paragraph{Zero-shot video classification.}

We use "CoCa" to denote CoCa image-text model that is directly applied to zero-shot video classification using frame-level average pooling as the feature aggregation. We use notation \textit{"w.o. videos"} to denote VideoCoCa model that is initialized from pretrained CoCa model without continued pretraining on VideoCC3M. On Kinetics 700, "Avg. Top 1/5" refers to averaged top-1 and top-5 accuracy, following ~\cite{alayrac2022flamingo}. 

Table \ref{tab:sota_cls} summarizes our zero-shot video classification results on Kinetics-400/600/700, UCF101, HMDB51, and Charades. On Kinetics, we observe that CoCa image-text model alone is already a strong baseline, leading to a $+2.5\%$  top-1 accuracy on Kinetics 400 and a $+1.6\%$ averaged top-1 and top-5 accuracy on Kinetics 700 compared to the previous state-of-the-art models. Compared to CoCa, We surprisingly find that VideoCoCa (\textit{w.o. videos}) with the plugin attentional poolers as the video adaptor is instantly adaptable to the flattened frame embeddings, yielding significant improvement on top of CoCa. Furthermore, when continually pretrained on VideoCC3M, VideoCoCa can be further improved, leading to $+5.6\%$, $+5.0\%$, $+2.3\%$ top-1 accuracy on Kinetics 400, Kinetics 600 and Kinetics 700 over the CoCa baseline. On UCF101 and HMDB51, CoCa can already achieve competitive results, mostly due to less human action variations in the video clip~\cite{kay_arxiv_2017}. VideoCoCa is again able to give further improvements on top of CoCa. On the long-form video dataset Charades, VideoCoCa achieves state-of-the-art results ($+4.7$ mAP) with only $8$ frames.

\vspace{-4mm}
\paragraph{Zero-shot text-video retrieval.}

We summarize our results on text-to-video retrieval in Table \ref{tab:sota_ret}. We also report video-to-text retrieval results following ~\cite{zeng2022socratic}. On MSR-VTT, ActivityNet Captions and VATEX, VideoCoCa achieves the highest zero-shot retrieval metrics for both text-to-video and video-to-text retrieval. We find that continued training on VideoCC3M is able to further improve on top of VideoCoCa (\textit{w.o. videos}), for example, by $+1.4$ R@1 on MSR-VTT text-to-video retrieval, $+2.3$ R@1 on ActivityNet Captions and $+2.5$ R@1 on VATEX. We observe even bigger gains on video-to-text retrieval. On Youcook2, VideoCoCa does not achieve better results. As we we observed in Table \ref{tab:mix_ablations}, increasing the ratio of HowTo100M can improve the results on Youcook2 but decrease the performance on MSR-VTT and ActivityNet Captions. This explains works such as ~\cite{yang2021taco,xu2021videoclip}, which are solely pretrained on HowTo100M, have strong results on Youcook2 but underperformed results on MSR-VTT. Similar findings are also reported in ~\cite{castro2022fitclip}. We conclude that the improvement on Youcook2 can be attributed to the strong visual-language domain match between YouCook2 and HowTo100M.  %Similar findings including their underperformed results on MSR-VTT are reported in \cite{castro2022fitclip}. 

\paragraph{Video captioning.}

We report both zero-shot and finetuned video captioning results in Table~\ref{tab:sota_cap}.  VideoCoCa significantly outperforms other models on zero-shot video captioning on MSR-VTT, ActivityNet Captions and VATEX. On Youcook2, we report results without using transcripts in the encoder for a fair comparison. For finetuning, VideoCoCa achieves competitive results among previous works, in some cases outperform them by large margins (\emph{e.g.} ActivityNet Captions, VATEX).

%VideoCoCa significantly outperforms AVR~\cite{nagrani2022learning} and CoCa. We also explored lightweight finetuning on top of VideoCoCa. Table \ref{tab:sota_cap} shows results on finetuning for video captioning. On Youcook2, we report results without using transcripts in the encoder for a fair comparison. On all downstream datasets, VideoCoCa is able to achieve competitive results among state-of-the-art models. 

%\vspace{2mm}
\paragraph{Finetuning for video question answering.}

Table \ref{tab:sota_qa} lists our results on open-ended VideoQA. VideoCoCa, again, achieves strong performance among state-of-the-art models. We also note that there is no single video-text model that performs significantly better than the others on all four video-text tasks.

\section{Conclusion}

In this work, we presented VideoCoCa, a simplified approach to transfer pretrained image-text CoCa models to video-text tasks in a zero-shot manner. Without any video-only or video-text data, VideoCoCa zero-shot transfer baseline already achieves state-of-the-art results on zero-shot video classification and zero-shot text-to-video retrieval. Furthermore, we investigated efficient finetuning on video-text data with VideoCoCa and achieved strong results on video captioning and video question answering. We conducted extensive ablations on other architecture options and tuning methods on a wide range of video-text tasks. We hope our proposed approach and empirical results can establish a strong video-text baseline on leveraging pretrained image-text foundation models for video-text tasks.

%and will further investigate training on large-scale video-text data from scratch.

%and adapting our model to other modalities such as audio for audio classification.

%on leveraging pretrained image-text foundation models for video-text tasks.

% and will investigate if our model can be further improved on video understanding tasks with a stronger text decoder such as  Chinchilla~\cite{hoffmann2022training} and PaLM~\cite{chowdhery2022palm}.

\paragraph{Acknowledgements.}
We thank Anja Hauth, Arsha Nagrani, David Ross, Bryan Seybold, Xuehan Xiong, Lucas Smaira, AJ Piergiovanni, Anelia Angelova, Radu Soricut, Yin Cui, Rui Qian, Antoine Yang, Mojtaba Seyedhosseini, Paul Natsev, Cordelia Schmid and Tom Duerig for their valuable feedback on dataset evaluation. We also appreciate Vijay Vasudevan for the helpful discussions on embedding extractions. We thank Xingyi Zhou, Wei Han and Claire Cui for their proofreading efforts, Erica Moreira, Victor Gomes for their assistance with resource coordination. Our implementation was made possible thanks to the support of the infrastructures team, and we want to specially thank Laurent El Shafey for his guidance on the infrastructures and model scalings.

%%%%%%%%% REFERENCES

{\small
\bibliographystyle{ieee_fullname}
\bibliography{egbib}
}

\twocolumn[
\centering
\Large
\vspace{1.0em}
] %
\appendix

\section{Varying the number of input frames.} \label{sec:scaling_num_frames}

The ablations on the zero-shot transfer performance with the increasing number of frames are shown in Table ~\ref{tab:varying_frames}. We evaluate on the Kinetics-400 and MSR-VTT datasets and report zero-shot video classification accuracies and text-to-video retrieval R@K. Results are obtained by VideoCoCa-Small and VideoCoCa-Base without continued pretraining on VideoCC3M. The image resolution $224\times224$ and patch size $16\times16$ is used. We observe that the performance gain on both video classification and text-to-video retrieval becomes smaller starting from 8 frames, and this is consistent across two model scales. Therefore, we keep 8 frames for each video clip to make the larger model trained faster and save more resources.

\section{Finetuning hyperparameters.} \label{sec: hparams}

Table~\ref{tab:training_hyperparameters} details the hyperparamters used in all of our VideoCoCa finetuning experiments for video captioning and video question answering.
We use Adafactor~\cite{shazeer2018adafactor} optimizer with $\beta_{1} = 0.9$,  $\beta_{2} = 0.99$ and decoupled weight decay ratio of 1e-4, a cosine learning rate schedule with linear warmup, and a batch size of $64$ or $128$ for our experiments. We precompute the frame embeddings with only center cropping applied on frames. We do not use other data augmentation methods (\emph{e.g.} random flip, color jittering, random augment~\cite{cubuk_arxiv_2019}) and other regularizations (\emph{e.g.} mixup~\cite{zhang_mixup_iclr_2018}, label smoothing~\cite{szegedy_cvpr_2016}, stochastic layer drop~\cite{huang_stochasticdepth_eccv_2016}). For video question answering, we additionally learn a fully connected layer (with identity activation) optimized with Softmax loss, and the learning rate of the classifier is set to be the same as the base learning rate for the decoder and the poolers.

\begin{table}[t]
\caption{Comparisons of zero-shot transfer performance with varying the number of frames on Kinetics-400 and MSR-VTT.
\label{tab:varying_frames}}
\centering
\small
\setlength{\tabcolsep}{4pt}
\aboverulesep=0ex
\belowrulesep=0ex
\resizebox{0.42\textwidth}{!}{%
\renewcommand\arraystretch{1.05}
\begin{tabular}{l|c|cc|ccc}
	\toprule 
 \multirow{2}{*}{\textbf{Model}} &
 %\multirow{2}{*}{\textbf{Num. Frames}} 
 \multirow{2}{*}{\makecell{\textbf{Num. Frames} \\ (\footnotesize{\(T\)})}}

 & \multicolumn{2}{c|}{\textbf{VideoCls.}} & \multicolumn{3}{c}{\textbf{VideoRet.}} \\
 &
 & \multirow{1}{*}{\footnotesize{Top-1}} & \multirow{1}{*}{\footnotesize{Top-5}} 
 & \multirow{1}{*}{\footnotesize{R@1}} & \multirow{1}{*}{\footnotesize{R@5}} 
 & \multirow{1}{*}{\footnotesize{R@10}} 
 \\
 \midrule

\multirow{7}{*}{\makecell{VideoCoCa-S \\ (\footnotesize{\textit{w.o. videos}})}}
& 1 & 33.1 & 59.2 & 13.5 & 29.3 & 37.6   \\ 
& 2 & 38.5 & 66.0 & 16.1 & 34.6 & 44.0  \\ 
& 4 & 42.1 & 70.4 & 21.7 & 42.8 & 52.8  \\ 
& 8 & 43.5 & 71.8 & 23.6 & 45.2 & 55.2  \\ 
& 16 & 44.2 & 72.3 & 24.3 & 45.5 & 55.7  \\ 
& 32 & 44.2 & 72.3 & 24.0 & 45.7 & 55.8  \\ 
& 64 & 44.3 & 72.4 & 24.1 & 45.6 & 55.6  \\ 
& 128 & 44.2 & 72.4 & 24.0 & 45.5 & 55.6  \\ 
 \midrule
\multirow{7}{*}{\makecell{VideoCoCa-B \\ (\footnotesize{\textit{w.o videos}})}}
& 1 & 49.4 & 72.8 & 18.5 & 36.5 & 44.9   \\ 
& 2 & 55.7 & 79.1 & 21.5 & 42.2 & 51.6  \\ 
& 4 & 59.3 & 82.5 & 24.7 & 47.3 & 57.4  \\ 
& 8 & 60.7 & 83.4 & 30.3 & 53.3 & 62.5  \\ 
& 16 & 61.1 & 83.8 & 30.7 & 53.6 & 62.7  \\ 
& 32 & 61.0 & 83.7 & 30.5 & 53.6 & 62.7  \\ 
& 64 & 61.1 & 83.8 & 30.5 & 53.6 & 62.6  \\ 
%& 128 & todo & todo & 30.4 & 53.4 & 62.4  \\ 
 \bottomrule
\end{tabular}
}
\end{table}

\begin{table*}[t]
\caption{Finetuning hyperparamters of VideoCoCa for video captioning and video question answering. We only use the center cropping for each video frame. We do not use other data augmentation methods (\emph{e.g.} random flip, color jittering, random augment) and other regularizations (\emph{e.g.} mixup, label smoothing, stochastic layer drop). 
Values which are constant across all columns are listed once.
Datasets are denoted as follows: MSRVTT, ActivityNet-Captions, VATEX and YouCook2 for video captioning, IVQA, MSRVTT-QA, MSVD-QA, and ActivityNet-QA for video question answering.}
\footnotesize{
\begin{tabularx}{\textwidth}{lcccccccc} %
\toprule
 & MSRVTT & ActivityNet-Captions & VATEX & YouCook2 & IVQA & MSRVTT-QA & MSVD-QA & ActivityNet-QA \\ \midrule
%\multicolumn{7}{l}{\textit{Optimization}}    \\
Optimizer               & \multicolumn{8}{c}{Adafactor with decoupled weight decay}  \\
Gradient clip               & \multicolumn{8}{c}{1.0}  \\
EMA decay rate					& \multicolumn{8}{c}{0.99} \\
Weight decay rate & \multicolumn{8}{c}{1e-4} \\
Learning rate schedule	& \multicolumn{8}{c}{Cosine with linear warmup} \\
Linear warmup epochs	& \multicolumn{8}{c}{1000} \\
Batch size & 64 & 64 & 64 & 64 & 64 & 128 & 128 & 128\\
Base learning rate	&	1e-5	& 1e-5 & 1e-5 & 1e-5 & 5e-6 & 1e-5 & 1e-5 & 1e-5 \\
Classifier learning rate & -- & -- & -- & -- & 5e-6 & 1e-5 & 1e-5 & 1e-5 \\
Training steps	 & 5000 & 5000 & 10000 & 10000 & 10000 & 10000 & 10000 & 10000 \\
\bottomrule
\end{tabularx}
}
\label{tab:training_hyperparameters}
\end{table*}

%\clearpage

\end{document}